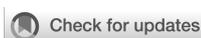

# A simplified retriever to improve accuracy of phenotype normalizations by large language models

Daniel B. Hier[1]*, Thanh Son Do[2] and Tayo Obafemi-Ajayi[3]

[1]Department of Neurology and Rehabilitation, University of Illinois at Chicago, Chicago, IL, United States, [2]Department of Computer Science, Missouri State University, Springfield, MO, United States, [3]Engineering Program, Missouri State University, Springfield, MO, United States

Large language models have shown improved accuracy in phenotype term normalization tasks when augmented with retrievers that suggest candidate normalizations based on term definitions. In this work, we introduce a simplified retriever that enhances large language model accuracy by searching the Human Phenotype Ontology (HPO) for candidate matches using contextual word embeddings from BioBERT without the need for explicit term definitions. Testing this method on terms derived from the clinical synopses of Online Mendelian Inheritance in Man (OMIM®), we demonstrate that the normalization accuracy of GPT-4o increases from a baseline of 62% without augmentation to 85% with retriever augmentation. This approach is potentially generalizable to other biomedical term normalization tasks and offers an efficient alternative to more complex retrieval methods.

KEYWORDS

phenotype normalization, large language model, small language model, cosine similarity, HPO, OMIM, retrievalaugmented generation

# Introduction

Large pre-trained language models are increasingly used in healthcare care, showing promise in performing a variety of complex natural language processing (NLP) tasks, such as text summarization, concept recognition, and answer questions (1–3). Large language models can identify medical concepts in the text and normalize them to an ontology (4–7). However, when large language models normalize medical terms to a standard ontology such as the human phenotype ontology (HPO), the retrieved code is not always accurate.

Shlyk et al. (8) demonstrated that the accuracy of large language models in term normalization tasks can be improved with Retrieval-Augmented Entity Linking (REAL). This method generates definitions of HPO terms and target terms needing normalization. The definitions are converted into word embeddings, and cosine similarity is used to identify the three closest candidate terms. The large language model then selects the best normalization from these candidates.

In this work, we introduce a simplified but effective retriever that bypasses the need for definition generation. Instead, it matches HPO terms to target terms using BioBERT contextual word embeddings, identifying the closest matches by semantic similarity. By prompting GPT-4o with the 20 closest candidate terms, we enable the model to take advantage of its implicit knowledge of HPO terms and select a semantically equivalent normalization. This approach achieves accuracy comparable to more complex methods without using explicit definitions.

Phenotyping, which involves recognizing the signs and symptoms of disease in patients and mapping them to an appropriate ontology such as HPO, is critical for precision medicine (9–13). Manual phenotyping is labor intensive, which makes high-throughput automated methods essential (11, 14–18). Phenotyping can be seen as part of the broader task of term normalization in different vocabularies, such as drugs to RXNORM (19), diseases to ICD-11 (20), or laboratory tests to LOINC (21).

A distinction can be made between surface phenotyping, which assigns a diagnosis from a disease ontology (22), and deep phenotyping, which assigns an HPO concept and ID to each symptom (11). Our work focuses on deep phenotyping (17).

Although often performed together, concept extraction (identification) and concept normalization are distinct NLP tasks. In concept extraction, the goal is to find relevant medical concepts within free text. Concept normalization involves matching irregular medical terms to standardized terms in an ontology and their corresponding machine-codes. If the irregular term to normalize has an exact match in the ontology, the task is straightforward such a matching "hyporeflexia" to its standard form in the HPO which is "Hyporeflexia." If the irregular term has no exact match (e.g., "diminished reflexes"), the model must select the most semantically similar concept (e.g., "decreased reflexes"). Additionally, concept normalization involves mapping irregular terms to their standard forms in an ontology and their corresponding machine codes ("hyporeflexia," HP:0001265).

Advances in deep learning, transformer architectures, and dynamic word embeddings have facilitated the development of tools that perform concept identification and normalization such as Doc2Hpo, ClinPheno, BERN2, PhenoBERT, and FastHPOCR (23–30). Evaluations of concept recognition tools show F1 values ranging from 0.50 to 0.76, with FastHPOCR performing the best on manually annotated corpora (29, 31).

Although deep learning and transformer-based methods have shown promise, they require extensive training data, which can be time consuming to acquire. Pre-trained large language models offer an alternative approach by eliminating the need for new training data. Despite their impressive performance, large language models may still make errors in retrieving the correct HPO ID (32).

Retrieval-augmented generation (RAG) (33) addresses this problem by using a retriever to provide relevant information, improving the likelihood of generating accurate output (34). In this paper, we extend the work of Shlyk et al. (8) by demonstrating that a simplified retriever, which relies on embedded terms, can improve normalization accuracy without the need to generate embedded term definitions. By prompting a large language model with candidate terms that have similarity to the target term, we achieve high accuracy with a more efficient approach.

## Methods

### Experimental plan

We selected 1,820 phenotypic terms from the Clinical Features sections of OMIM summaries as a test set for term normalization. In the first experimental condition, the NLP models (spaCy and BioBERT) normalized the terms by selecting the best-matching HPO term and HPO ID based on the cosine similarity of the embedded word vectors. In the second experimental condition, language models were prompted to normalize a term via the OpenAI API to the best matching HPO term and HPO ID. In the third experimental condition, the prompts for the language models were augmented with up to 50 candidate terms and HPO IDs generated based on the cosine similarity between the BioBERT word embeddings and the term to be normalized. For each experimental condition, we calculated accuracy, F1, recall, and precision of term normalization.

### Data

Terms to normalize were the signs and symptoms of neurogenetic diseases derived from *Clinical Feature* summaries in the OMIM database (35). We downloaded clinical feature summaries for 236 neurogenetic diseases, consisting of 175,724 tokens via the OMIM API (https://api.omim.org). GPT-3.5-Turbo was used to identify 2,023 terms for normalization (mean 16.5 signs per disease) (18, 36). The text for extracting signs and symptoms was passed to the GPT-3.5 Turbo API with the following prompt:

```
prompt =
(You are a neurologist analyzing a case summary.
The input is a JSON object containing:
"clinical Features"
Your task is to extract all relevant
neurological symptoms (patient complaints) and
signs (findings on examination).
Exclude any signs and symptoms related to family
members.
Please respond with the findings organized into
a dictionary under the key "Signs."
Each sign should be distinctly listed.
Here is the format for your response:
{
"Signs": ["sign a," "sign b," "sign c"]
}
Report only signs and symptoms observable by
the physician at the bedside.
Ignore all laboratory, pathological,
and radiological signs)
```

A domain expert excluded 203 malformed terms (e.g., vague, contradictory, verbose, or ambiguous phrases). These terms were excluded because they were judged to be difficult to normalize. Examples of malformed terms that would be difficult to normalize included:

```
impaired visual pathways
initial good response to dopaminergic therapy
intermittent microsaccadic pursuits
intermittent mobility
intermittent tetanic contraction
intrusive square wave jerks
jerky voice
kineto rigid syndrome
legs and arms
```

The final test dataset consisted of 1,820 terms to normalize.

The Human Phenotype Ontology (HPO) was downloaded as a comma-separated value (CSV) file from NCBO BioPortal (37). A list of 17,957 HPO entry terms was expanded to 30,234 by adding all available synonyms. Each HPO entry term was associated with a corresponding HPO ID, formatted as `HP:nnnnnnn`, where `n` is a digit between 0 and 9.

## Term normalization using NLP-based methods

**spaCy**: SpaCy was combined with `en_core_web_lg` word embeddings. Vectors were generated for each HPO entry term and stored as a Python dictionary.

**BioBERT**: We utilized the BioBERT v1.1 model (`dmis-lab/biobert-base-cased-v1.1`) to compute embeddings for target terms and HPO entry terms (38). Each term (target or HPO term) was tokenized using the BioBERT tokenizer. The resulting embeddings were computed using the BioBERT transformer model, and the mean of the token embeddings across all tokens was used as the global embedding vector:

```
inputs = tokenizer(term, return_tensors="pt,"
truncation=True, padding=True, max_length=128)
with torch.no_grad():
outputs = model(**inputs)
embedding = outputs.last_hidden_state.mean
(dim=1).squeeze().numpy()
```

HPO entry terms and their corresponding IDs were preprocessed, and their embeddings were calculated and stored in a CSV file for efficiency. For each target term, cosine similarity between its BioBERT embedding and the precomputed HPO embeddings was calculated. The HPO term with the highest similarity score was selected as the "best match":

```
similarities = cosine_similarity(term_vector,
hpo_embeddings).flatten()
best_match_idx = np.argmax(similarities)
```

The same method was used to retrieve the **k** best matches for inputting candidate terms to the the large language model with retriever methods.

**Doc2Hpo**: Terms were normalized using the Doc2Hpo API at https://doc2hpo.wglab.org/parse/acdat using the string-based matching engine (25).

## Term normalization using large language models

Three large language models were evaluated for term normalization: GPT-4o, GPT-3.5-Turbo, and GPT-4o-mini from OpenAI (San Francisco, CA) via its API (https://api.openai.com). Each of the terms to normalize was passed to the API with the following prompt:

```
prompt = (
You are given a term to normalize to a concept
from the Human Phenotype Ontology and return
the best match and its HPO ID.
"Term: {term}"
Pick the best one and return it in JSON format:
{"best_match": "term," "HPO ID": "HP:nnnnnnn"}
```

## Term normalization using large language models enhanced with retrieval augmentation

The performance of large language models for term normalization was enhanced by augmenting the prompt with up to 50 candidate terms. The top 20 candidates from the BioBERT embeddings were used in the final analysis.

```
prompt = (
You are given a term to normalize to a concept from
the Human Phenotype Ontology and its HPO_ID:
Term: {term}
Possible matches: [match_1...match_20]
Pick the best one from the above matches and
return it in JSON format:
{"best_match": "term," "hpo_id": "HP:xxxxxxx"}
```

## Assessment of semantic equivalence

We evaluated the semantic equivalence of the normalized terms by comparing the "best matches" to the original *terms to normalize*. Among the 1,820 terms, 438 had exact matches in the list of HPO entry terms. To assess semantic equivalence, we employed three complementary approaches:

1. **Cosine Similarity:** We calculated the cosine similarity between the embeddings of the term to normalize and the candidate HPO terms using BioBERT embeddings.
2. **GPT-3.5-Turbo Judgment:** GPT-3.5-Turbo was prompted to assess semantic equivalence by returning a binary judgment (equivalent or not) for each input term.
3. **Expert Review:** Domain experts in clinical terminology provided the final judgment on semantic equivalence, taking into account the cosine similarity and the GPT-3.5-Turbo binary judgment.

A term was deemed an accurate match (True Positive, TP) if it was semantically equivalent to the original term and was mapped to the correct HPO ID. A False Positive (FP) occurred when a term was semantically incorrect or the HPO ID was inaccurate. Malformed terms that were previously excluded from normalization attempts were not counted as True Negatives (TN). If a model failed to return any normalization for a given term, it was considered a False Negative (FN). Accuracy, F1, recall, and precision were calculated using standard formulas:

Accuracy = (TP + TN) / (TP + TN + FP + FN)
Precision = TP / (TP + FP)
Recall = TP / (TP + FN)
F1 = 2 × (Precision × Recall) / (Precision + Recall)

Metrics were reported to two decimal places, reflecting the precision and reproducibility of the calculations.

# Results

Table 1 shows the model accuracies for the phenotype normalization of the 1,820 *terms to normalize*. To be rated as "accurate," the *term to normalize* had to be semantically equivalent to the HPO entry term and and the model had to retrieve the term's correct HPO ID.

The spaCy and BioBERT methods used word embeddings and NLP algorithms to find the best match in a complete table of HPO terms. The spaCy embeddings were general-purpose word embeddings, whereas the BioBERT embeddings were optimized for biomedical terminologies. spaCy averaged the vectors of the component tokens to get a global term vector, whereas BioBERT utilized a transformer architecture and hidden states to generate a global term vector. Both methods used cosine similarities to find the best match for each *term to normalize* and candidate terms in the HPO. The BioBERT method, at 69% accuracy, outperformed the spaCy method at 46%.

The GPT-4o-mini, GPT-3.5-Turbo, and GPT-4o models without a retriever had no access to external data sources and relied entirely on pre-training to find HPO IDs. Errors made by these language models typically involved the retrieval of incorrect HPO IDs rather than errors in the HPO entry terms. In most cases, when the models made an error, the HPO entry term was correct or nearly correct, but the HPO ID was inaccurate and matched an incorrect concept in the HPO. Among large language models without a retriever, GPT-4o, the largest and most advanced model, performed best with a accuracy of 62%, while GPT-4o-mini, the smallest model, performed worse with an accuracy of 12%.

The best-performing methods combined a language model with a retriever. Each model had 20 candidate normalizations for a *term to normalize* based on the closest embedding similarities. Each language model was prompted to choose the "best match" from the twenty closest candidates. In this scenario, the GPT-4o and GPT-3.5-Turbo models outperformed the BioBERT method with 85% to 88% accuracies (Table 1).

We investigated the optimal number of potential matches to submit to GPT-4o or GPT-3.5-Turbo (Figure 1). Accuracy improved as the number of candidate normalization increased to 20 and then reached a plateau with further increases in candidates not improving accuracy of normalization.

# Discussion

The results indicate that the most accurate method for phenotype term normalization combines a language model with a retriever. GPT-4o and GPT-3.5-Turbo, when paired with a retriever, achieved the highest accuracies of 85% to 88%, demonstrating the benefits of augmenting language models with a retrieval mechanism.

The standalone BioBERT, specifically optimized for biomedical text, performed better than spaCy or GPT-4o without retrieval, with an accuracy of 69%. This highlights the limitations of a standalone large language model relying solely on pre-training when no external retrieval is available. BioBERT's ability to generate specialized biomedical embeddings allowed it to outperform GPT-3.5-Turbo and GPT-4o without a retriever.

GPT-4o-mini, smaller and less pre-trained than GPT-4o and GPT-3.5-Turbo, showed weak performance on term normalization with 12% accuracy, highlighting its lack of exposure to HPO terms and their HPO IDs during training.

Examining the cases where large language model combined with a retriever outperformed BioBERT reveals that the "best match" selected by large language model can deviate from the candidate term with the highest cosine similarity (Table 2). This demonstrates the strength of large language models in interpreting semantic equivalence beyond simple cosine metrics. For example, "foot drop" was selected as a better semantic match for "bilateral foot drop" than "bilateral clubfoot," despite having a lower cosine similarity score. Similarly, "depigmented fundus" was preferred for "pale fundi" over "pale eyelashes." The ability of a large language model to override cosine similarity based on contextual understanding explains the superior accuracy of the retriever-augmented method. Part of this superiority likely resides on focusing attention on the semantically most important words in compound terms such as "depigmented fundu" where the focus is appropriately directed to "fundus" in favor of the less important word "depigmented."

Our results suggest that, while large language models possess significant capabilities, their performance on phenotype normalization tasks can be enhanced by retrieval augmentation.

TABLE 1 Model metrics for term normalization to HPO.

| Method | Accuracy | F1 | Recall | Precision | N |
| --- | --- | --- | --- | --- | --- |
| spaCy embeddings cosine similarity | 0.46 | 0.63 | 0.46 | 1.00 | 1,820 |
| BioBERT embeddings by cosine similarity | 0.69 | 0.81 | 0.69 | 1.00 | 1,820 |
| GPT-4o mini | 0.12 | 0.21 | 0.40 | 0.40 | 1,820 |
| GPT-3.5-Turbo | 0.51 | 0.67 | 1.00 | 0.51 | 1,820 |
| GPT-4o | 0.62 | 0.77 | 0.95 | 0.65 | 1,820 |
| Doc2Hpo API | 0.63 | 0.77 | 0.62 | 0.99 | 1,820 |
| GPT-3.5-Turbo with retrieval augmentation | **0.88** | 0.93 | 0.96 | 0.91 | 1,820 |
| GPT-4o with retrieval augmentation | **0.85** | 0.92 | 0.96 | 0.88 | 1,820 |

Note: The sample size (N) for all methods is 1820. For augmented methods, the language models were presented with a set of 20 candidate terms generated by the retriever. False negatives (FN) were excluded as "malformed terms" (see Methods).
Bold values show models with highest accuracy on term normalization task.

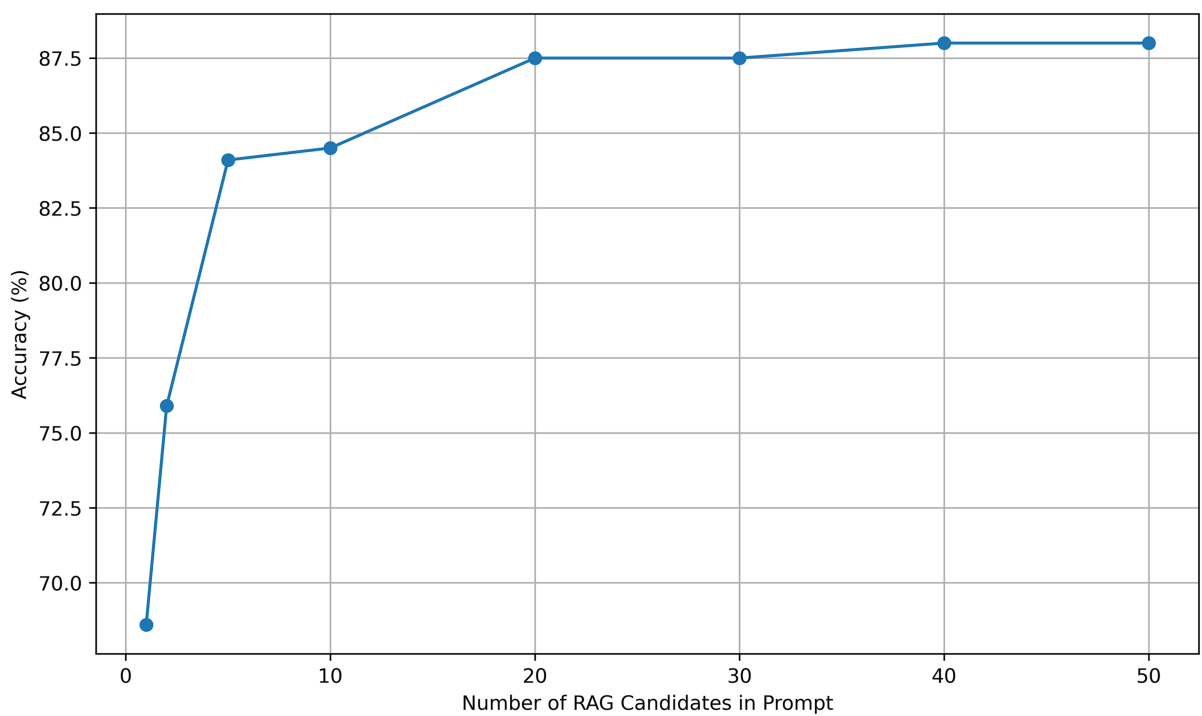

FIGURE 1
RAG candidate choices vs. accuracy. Accuracy was evaluated for normalization by GPT-3.5-Turbo with candidate prompts in the range of 1 to 50. Accuracy improved until reaching a plateau at 20 candidate terms in the prompt.

The accuracy of normalization improves when the large language model is presented with candidate terms selected via cosine similarity. The plateau observed for 20 candidate terms in the prompt suggests that presenting more candidates neither degrades nor improves accuracy (Figure 1).

Shlyk et al. (8) demonstrated the efficacy of retrieval-augmented entity linking using embedded term definitions. However, our approach, which bypasses the need for explicit definitions, achieves comparable results. This suggests that advanced models like GPT-4o and GPT-3.5-Turbo can draw on pre-trained knowledge to assess semantic equivalence without relying on external definitions, streamlining the normalization process.

Regarding limitations, our study focused solely on term normalization and did not evaluate term identification. Additionally, our definition of "semantic equivalence" remains qualitative rather than exact. Determining whether "impaired sensation" is semantically equivalent to "decreased sensation" can be approached in three ways: (1) by measuring cosine similarity between appropriate word embeddings, (2) by the judgment of large language models trained for semantic reasoning, or (3) by gold-standard review from human domain experts. However, as datasets grow beyond 2,000 terms, human review becomes increasingly impractical, particularly for large ontologies such as SNOMED CT with over 400,000 terms. Another limitation is the relatively small and specialized list of terns derived from the OMIM summaries of neurogenetic diseases, which may not capture the full diversity of phenotype terms that require normalization. Expanding the dataset to cover a broader range of phenotypic terms from other domains could provide further insight.

TABLE 2 GPT-3.5-Turbo may select an HPO term as semantically equivalent that did not have the highest cosine similarity by BioBERT word embeddings.

| Term to normalize | BioBERT match by CS | CS | Large language model + retriever match | CS | Δ |
| --- | --- | --- | --- | --- | --- |
| Absent ankle jerks | Absent knee jerk reflex | 0.96 | Absent ankle reflexes | 0.95 | 0.01 |
| Pale fundi | Pale eyelashes | 0.92 | Depigmented fundus | 0.91 | 0.01 |
| Lack of speech | Poor speech discrimination | 0.93 | Absent speech development | 0.92 | 0.01 |
| Disinhibition | Inactivity | 0.89 | Social disinhibition | 0.88 | 0.01 |
| Hand weakness | Shoulder weakness | 0.97 | Hand muscle weakness | 0.96 | 0.01 |
| Bilateral foot drop | Bilateral clubfoot | 0.94 | Foot drop | 0.93 | 0.01 |

Note: The table shows the cosine similarity (CS) between the "term to normalize" and the term choice by the BioBERT method and the term choice by the large language model + Retriever. Using the BioBERT method, the term in the HPO with maximal CS to the "term to normalize" is chosen. The large language model + Retriever method selects from 20 candidate terms the the term with best "semantic equivalence" while ignoring cosine similarity. As shown in this Table, in some cases large language model + Retriever can outperform the BioBERT method due to its ability to pick terms that are better matches to the "term to normalize" that do not have the highest cosine similarities. Δ is the difference between the cosine similarities for the term selected for each method.

Looking ahead, our retriever-based method is potentially generalizable to other terminologies, such as Gene Ontology (GO), UniProt, and SNOMED CT. By avoiding term definitions and relying solely on word embeddings, our approach could simplify normalization tasks in domains where definitions are ambiguous or difficult to generate. Further research into alternative retrieval strategies could improve the accuracy of the model and expand its applicability.

In conclusion, retrieval-augmented prompts based on BioBERT word embeddings improve the accuracy of phenotype normalization tasks. This simplified retriever performs as well as methods based on term definitions, offering an alternative solution for biomedical term normalization.

# Data availability statement

Publicly available datasets were analyzed in this study. The Human Phenotype Ontology (HPO), which includes 22,929 classes, is available for download as a CSV file from the National Center for Biomedical Ontology at https://bioportal.bioontology.org/ontologies/HP. A dataset containing 271,702 HPO annotations for 8,259 diseases in OMIM and 4,283 diseases in Orphanet can be accessed at https://hpo.jax.org/data/annotations. The Python code used in this project is publicly available on the project's GitHub repository at https://github.com/clslabMSU/simplified-retriever.

# Ethics statement

Ethical approval was not required for the study involving humans in accordance with the local legislation and institutional requirements. Written informed consent to participate in this study was not required from the participants or the participants' legal guardians/next of kin in accordance with the national legislation and the institutional requirements.

# Author contributions

DBH: Conceptualization, data curation, formal analysis, methodology, writing – original draft, writing – review & editing. TSD: Formal analysis, validation, writing – original draft, writing – review & editing. TO-A: Formal analysis, methodology, validation, writing – original draft, writing – review & editing.

# Funding

The author(s) declare that no financial support was received for the research, authorship, and/or publication of this article.

# Conflict of interest

The authors declare that the research was conducted in the absence of any commercial or financial relationships that could be construed as a potential conflict of interest.

# Publisher's note

All claims expressed in this article are solely those of the authors and do not necessarily represent those of their affiliated organizations, or those of the publisher, the editors and the reviewers. Any product that may be evaluated in this article, or claim that may be made by its manufacturer, is not guaranteed or endorsed by the publisher.